\documentclass[letterpaper, 10 pt, conference]{ieeeconf}  

\IEEEoverridecommandlockouts                              

\usepackage{graphics} 
\usepackage{epsfig} 
\usepackage{mathptmx} 
\usepackage{times} 
\usepackage{cite}
\usepackage{amsmath,amssymb,amsfonts}
\usepackage{graphicx}
\usepackage{algorithmic}
\usepackage{multirow}
\usepackage[hyphens]{url}
\usepackage{textcomp}
\usepackage{color,soul}
\usepackage{booktabs}

\title{\LARGE \bf
HELP: Hierarchical Embodied Language Planner for Household Tasks
}

\author{Alexandr V. Korchemnyi$^{1}$, Anatoly O. Onishchenko$^{1}$, Eva A. Bakaeva$^{1}$,\\Alexey~K.~Kovalev$^{2,1}$, and Aleksandr~I.~Panov$^{2,1}$
\thanks{$^{1}$MIRAI, Moscow, Russia
        {\tt\small{(\{onishchenko.a, kovalev.a\}@miriai.org)}}, $^{2}$Cognitive AI Systems Lab, Moscow, Russia {\tt\small(panov@cogailab.com)}}%
}

\begin{document}

\maketitle
\thispagestyle{empty}
\pagestyle{empty}

\begin{abstract}
Embodied agents tasked with complex scenarios, whether in real or simulated environments, rely heavily on robust planning capabilities. When instructions are formulated in natural language, large language models (LLMs) equipped with extensive linguistic knowledge can play this role. However, to effectively exploit the ability of such models to handle linguistic ambiguity, to retrieve information from the environment, and to be based on the available skills of an agent, an appropriate architecture must be designed. We propose a Hierarchical Embodied Language Planner, called HELP, consisting of a set of LLM-based agents, each dedicated to solving a different subtask. We evaluate the proposed approach on a household task and perform real-world experiments with an embodied agent. We also focus on the use of open source LLMs with a relatively small number of parameters, to enable autonomous deployment.
\end{abstract}

\section{INTRODUCTION}
\label{sec:introduction}
One line of research in embodied artificial intelligence focuses on equipping agents, such as robots, with the ability to execute complex tasks by following natural language instructions~\cite{huang2022language,kovalev2022application,sarkisyan2023evaluation,onishchenkolookplangraph}. This approach facilitates human-robot interaction by eliminating the need for users to understand intricate control mechanisms. However, natural language understanding presents inherent challenges, including lexical and pragmatic ambiguities~\cite{knowno2023,ivanova2025ambik}, information gaps~\cite{padmakumar2022teach}, and coreference resolution issues~\cite{park2023clara}. To achieve greater autonomy and adaptability in complex environments, embodied agents must therefore be capable of planning and acting despite such ambiguous instructions.

For effective planning, whether in simulators~\cite{alfred, virtual_home,padmakumar2022teach,kachaev2025mind} or real-world environments~\cite{huang2022inner,saycan2022arxiv}, the planner must comprehend the preconditions and postconditions of each action~\cite{grigorev2024common,grigorev2025verifyllm}, as well as the affordances and limitations of the objects involved~\cite{chuganskaya2023problem}. This comprehension is necessary to construct a precise action sequence that leads to the successful completion of tasks.
Work~\cite{huang2022language} has demonstrated that pre-trained Large Language Models (LLMs) can assume this planning role without requiring task-specific fine-tuning.
\begin{figure}[t]
    \centering
    \includegraphics[width=\linewidth]{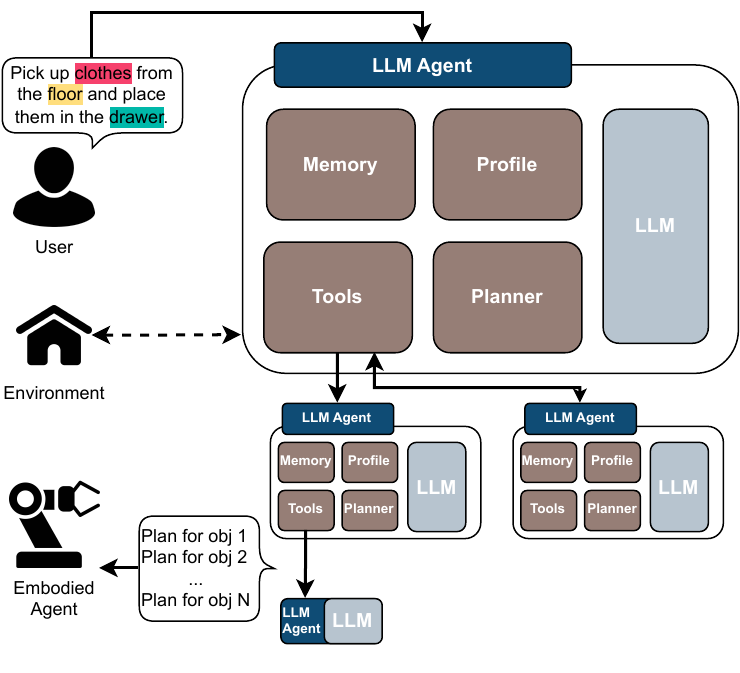}
    \caption{Given a high-level natural language instruction, a high-level LLM agent decomposes it into unambiguous subtasks (e.g., “Pick up the shirt,” “Pick up the jeans,” etc.). Each subtask is passed to a low-level LLM agent, which generates executable pseudocode grounded in the robot’s action space.}
    \label{fig:visual_abstract}
    \vspace{-10pt}
\end{figure}
LLMs can be used for planning by scoring potential actions~\cite{saycan2022arxiv} or by directly generating a sequence of steps in natural language~\cite{huang2022inner} or pseudocode~\cite{singh2023progprompt}. One approach utilizes a hierarchical framework (Figure~\ref{fig:visual_abstract}) in which an LLM is tasked with generating a high-level plan, as demonstrated in~\cite{song2023llm, palm-e}.
However, because LLMs rely on the extensive but static knowledge acquired from their training corpora, they tend to propose solutions that are generically plausible rather than specifically tailored to a unique situation. Consequently, a plan may fail if the actual state of the environment differs from the model's expectations~\cite{pchelintsev2025lera,grigorev2025verifyllm,yudin5370686scene,onishchenkolookplangraph}. This problem is exacerbated when natural language instructions from a human lack complete or clarifying information about the specific environment.
 
To illustrate, consider the task: ``Pick up the clothes from the floor and place them in the drawer.'' Successful execution requires recognizing that ``clothes'' refers to multiple distinct objects. Consequently, the agent must first identify these individual items and then formulate a specific plan for each one. This process creates a need for environmental feedback, which can be provided by transmitting a list of visible objects, embedding sensory observations, maintaining a history of completed actions, or assessing the feasibility of available actions.
In general, existing LLM-based planning methods show promising results but suffer from several downsides: (a) they demand significant computational resources, rendering them impractical for direct deployment on embodied agents; (b) they struggle to scale to environments with large numbers of objects or actions, due to problems such as context window overflow, hallucinations, and the high computational cost of scoring numerous potential actions.

\begin{figure*}[ht]
    \centering
    \includegraphics[width=\linewidth]{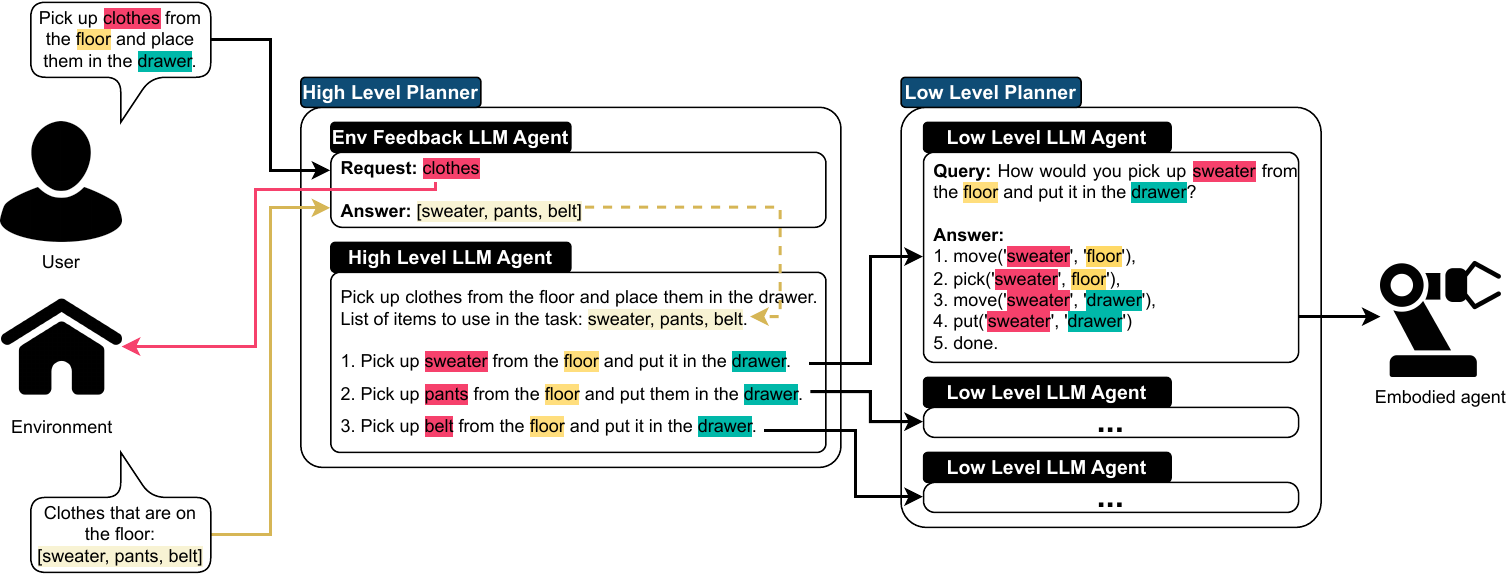}
    \caption{The HELP architecture consists of multiple LLM-based agents operating in a pipeline. The process begins with the High Level Planner (HLP), which resolves ambiguities by querying the environment for feedback if necessary. The HLP then decomposes the initial instruction into a set of individual natural language sub-tasks. These sub-tasks are subsequently passed to the Low Level Planner (LLP). The LLP translates each natural language sub-task into an executable plan, i.e., a sequence of actions grounded in the agent's available skills and the environment. Finally, the embodied agent executes the resulting sequence of actions.}
    \label{fig:method_overview}
\end{figure*}

We propose the \textbf{H}ierarchical \textbf{E}mbodied \textbf{L}anguage \textbf{P}lanner (HELP), a hierarchical LLM-based agent designed for medium-sized models (7-13B parameters). The planning process is divided into two distinct stages (Figure~\ref{fig:method_overview}). In the high-level planning stage, the initial natural language task is decomposed into a sequence of unambiguous natural language sub-tasks. These sub-tasks are then passed to the low level planner, which translates each one into a precise sequence of executable steps in pseudocode to guide the embodied agent. By logically splitting the problem, using natural language to resolve ambiguity and pseudocode to ensure precise grounding in the agent's available actions, we significantly reduce the complexity of the task for the language model. Consequently, our approach achieves robust performance with models of only 7-13 billion parameters, unlike prior methods that require models with 100 billion parameters or more~\cite{palm-e}. This efficiency enables planning directly on the embodied agent without relying on proprietary APIs.
Our methodology aligns with the paradigm of LLM-based multi-agent systems~\cite{wang2024survey}, where distinct LLM queries (agents) collaboratively solve a decomposed task. In our framework, the high level and low level planners act as autonomous agents. Their capabilities are augmented by: (a) a defined profile (via capability descriptions), (b) memory (short-term via in-context examples and long-term via multimodal maps and tracking), (c) tools (by delegating to other agents or querying memory), and (d) inherent planning capabilities. While such agent architectures are widely used in other fields, we specifically adapt and apply this approach to embodied robotics.

We validated HELP on a specially collected template-based dataset containing instructions with ambiguous references. Also, to demonstrate that our approach also works with unambiguous instructions, we evaluated it on the ALFRED dataset~\cite{alfred}. Furthermore, we embedded HELP into the existing control system of a real robotic platform and conducted experiments in which a human commanded the robot to perform a household item-sorting task via natural language instructions.

The contributions of this paper are as follows:
\begin{enumerate}
    \item We propose HELP, a hierarchical LLM-based planner for embodied agents that adheres to the multi-agent paradigm and is designed for extensibility to novel tasks.
    \item We validate HELP on a range of household tasks and demonstrate its practical efficacy by deploying it on a real-world embodied agent: a mobile platform equipped with a robotic arm.
    \item HELP is designed for medium-sized LLMs, enabling it to achieve strong performance on metrics for long-horizon planning. This efficiency allows the language model to be run directly on the embodied agent's hardware, eliminating dependence on large compute clusters or proprietary APIs.
\end{enumerate}

\section{RELATED WORK}
\label{sec:related-work}
Planning with LLMs is a relatively new and rapidly developing area. The foundations were laid in~\cite{huang2022language}, where it was proposed to use pre-trained LLMs and to ground the model-generated plan steps on available actions in the environment through the closeness of their text embeddings. In~\cite{saycan2022arxiv}, the LLM was not used in the generation mode, but it was evaluated which of the possible steps will lead to the final goal. In addition, an affordance function was used to evaluate actions in terms of their execution in the environment. These works generate steps in natural language and use low-level skills associated with a specific step formulation. A separate area of research is the generation of a plan in the form of program code~\cite{liang2023code,singh2023progprompt}, or the use of a special language for planning~\cite{liu2023llm+,silver2023generalized}, which on the one hand solves some problems with language ambiguity, and on the other hand requires LLMs that are good at generating program code.

A limitation of using LLM for planning in embodied tasks is that the LLM has a limited understanding of the environment, usually only through the description in the prompt, which can lead to the generation of impossible plans. Several works~\cite{huang2022language,saycan2022arxiv} solve this problem by modifying the generated plan through external mechanisms. However, a promising direction is to process feedback from the environment during plan generation or execution. In InnerMonologue~\cite{huang2022inner}, LLM can request additional information from the user or use information about the success of actions. ProgPrompt~\cite{singh2023progprompt} uses a hierarchical approach where one prompt is used to generate a plan and a second prompt is used to check the success of actions in the environment. LLM-Planner~\cite{song2023llm} provides an approach to solving the problem of mismatching generated and ground truth object names, but unlike in~\cite{huang2022language}, it implements this in the process of executing the plan, and not at the generation stage. In REFLECT~\cite{liu2023reflect}, the LLM is used to detect errors in the execution of the plan at the end of the episode, based on the summation of multimodal information from the environment.

Our work addresses the challenge of generating long-horizon plans by employing a hierarchical approach that decomposes the original task into atomic sub-tasks. To the best of our knowledge, this specific application of hierarchical decomposition for plan length management in embodied AI has not been previously explored. Furthermore, our method incorporates environmental feedback by querying object descriptions to disambiguate user instructions. Unlike prior work, which relies on massive proprietary models like GPT-3~\cite{brown2020language}, our approach is designed for open-source models~\cite{vicuna2023}with significantly fewer parameters. This focus on smaller, efficient models mitigates the computational barriers to deployment, enabling practical application on real robotic platforms without dependence on external APIs.

\section{PROBLEM STATEMENT}
\label{sec:problem}
Consider task planning for an embodied agent (a robot), denoted as $R = \langle S, P \rangle$, which is equipped with a set of skills $S$ and perceptual sensors $P$. The goals for the agent are defined by a set $G$ of feasible natural language instructions.

We formalize the task planning problem using a function $f: R \times G \rightarrow T$, where $T$ represents the set of all possible plans. A plan $p \in T$ is a sequence of skills, $p = (s_1, s_2, \dots, s_k)$ with $s_i \in S$, that leads to the achievement of a goal $g \in G$. The length of a plan $p$ is denoted by $|p| = k$.

Let $L$ be the maximum possible length of any plan needed to accomplish a goal from $G$ in a given environment; $L$ is assumed to be dependent on the environment and the specific instruction. Therefore, the codomain of the planning function can be considered as $S^{\leq L}$, the set of all skill sequences of length at most $L$.

Given that there may be multiple feasible plans for a goal, the optimal plan $\hat{p}$ is chosen to be the one with the shortest length, expressed as:
\begin{equation}
    \hat{p} = \arg\min\limits_{p \in T} |p| 
\end{equation}
subject to $p$ achieves $g$,where $T$ is the set of all feasible plans for goal $g$.

\section{PRELIMINARIES}
\label{sec:pre}
\subsection{Generative LLM}
The task of language generation for LLMs is formalized as follows: given a sequence of tokens \(X = (x_1, x_2, \ldots, x_{n-1})\), where each token \(x_i\) 
is an element from a fixed vocabulary $V$, the objective is to generate the next token $x_n$. This is achieved by selecting the token with the highest conditional probability:
\begin{equation}
    x_n = \arg\max\limits_{x \in V} P(x|X, \theta),
\end{equation}
where $\theta$ represents the model's parameters.

\subsection{Agent Architecture}
An LLM-based agent is a composite entity comprising an LLM and various auxiliary components that extend its capabilities. In our work, we follow the generalized architecture proposed in~\cite{wang2024survey,xi2023rise}. An LLM-based agent is formally represented as the tuple $A = \langle \theta, M, T, P, I \rangle$, where $\theta$ represents the parameters of the underlying language model. These parameters are typically shared across multiple agents. $M$ denotes the agent's memory, which consists of: 1) short-term memory, used for storing in-context examples, the current chain of thought~\cite{chain-of-thought}, or other transient information for immediate reasoning; 2) long-term memory, used for retaining persistent information such as environment maps, tracking data, multimodal observations, or histories of successfully completed trajectories. $T$ represents the set of tools at the agent's disposal. This can include external APIs, knowledge bases, databases, other specialized models, or functions internal to the agent itself. $P$ denotes the agent's planning module, which is responsible for decomposing high-level tasks into a sequence of executable steps. $I$ signifies the agent's profile, which defines its identity, role, and capabilities through a natural language description. This provides context for the LLM's operation. While our work does not utilize psychological or sociological features, this component is included for architectural completeness. It is important to note that a specific agent implementation may not utilize all of these components, allowing for flexibility within this architectural framework.

\subsection{High Level Planner}
The High Level Planner (HLP) is an instance of an LLM agent, formally defined as $H=\langle \theta, M^H, T^H, P^H, I^H \rangle$.
The memory component $M^H$ is instantiated as $M_e^H$, which provides contextual examples for in-context learning. The toolset is defined as $T^H = (F, L)$, where $F$ is a function (or an LLM agent) responsible for determining if a sub-task requires disambiguation and for querying long-term memory to retrieve relevant information; $L$ is the Low Level Planner agent. Planner $P^H$ encapsulates the mechanism for decomposing an initial high-level task, expressed in natural language, into a linear sequence of low-level natural language sub-tasks. This sequence is subsequently passed to $L$ for execution. Profile $I^H$ is a natural language description that defines the agent's role as an intelligent assistant and planner.

\subsection{Low Level Planner}
The Low Level Planner (LLP) is an LLM agent denoted by 
$L = \langle \theta, M^L, T^L, P^L, I^L \rangle$, where $M^L = M_e^L$ is its memory, which consists of a list of contextual examples. $T^L = {f_g^L}$ is its set of tools, where $f_g^L$ is a grounding function that maps the planner's textual output to the specific objects and actions available in the environment. $P^L$ is its planning mechanism, which decomposes a single natural language sub-task (provided by the HLP) into a sequence of executable actions for the robot. $I^L$ is its profile, which defines the agent's role and capabilities.

\subsection{Feedback Mechanisms}
An LLM agent, denoted as $F = \langle \theta, M^F, T^F, P^F, I^F \rangle$, is employed to query the environment for feedback. The memory is $M^F = (M_e^F, M_m^F)$, consisting of: 1) short-term memory $M_e^F$ (a list of contextual examples); 2) long-term memory $M_m^F$ (a knowledge store containing information about objects in the environment). This may take the form of a multimodal map or a current observation. Profile $I^F$ is a description defining the agent's role. Planner $P^F$ is a simple mechanism for formulating a query. The primary objective of agent $F$ is to discern the appropriate request to be directed toward the long-term memory based on the task description. For instance, given the task \emph{``Put all the clothes from the floor into the drawer,''} the agent would generate the query \emph{``clothes''}. This query is then passed to a tool $T^F$, which uses it to retrieve the information from $M_m^F$ and returns the result to the HLP.

\section{METHOD}
\label{sec:method}
We introduce the \textbf{H}ierarchical \textbf{E}mbodied \textbf{L}anguage \textbf{P}lanner (HELP), a method designed to enable an embodied agent to execute natural language instructions in a real home environment (Figure~\ref{fig:method_overview}). The core architecture of HELP is a two-stage planning pipeline:
\begin{enumerate}
    \item \textbf{High Level Planning:} In this initial phase, ambiguities in the instruction are resolved, additional information is queried from the environment if necessary, and the task is decomposed into a sequence of elementary sub-tasks expressed in natural language.
    \item \textbf{Low Level Planning:} In the second phase, each natural language sub-task is translated into an executable plan, a sequence of actions grounded in the agent's available skills and the specific objects present in the environment.
\end{enumerate}
Both the high-level and low level planners are implemented as LLM-based agents.

HELP takes natural language instructions as input. Example queries include an operation on a single object (e.g., \emph{Pick up a pillow from the floor and put it on the couch''}), on several objects (e.g., \emph{Pick up a bowl and put it on the table, and then place a spoon there as well''}), or on an indeterminate number of objects (e.g., \emph{``Put all the stuff from the floor into the closet''}). The model's output is a sequence of low-level actions generated by the LLM and presented in a structured text format. This sequence is designed for execution by the embodied agent. For the first example, the output might be:
\textit{1. move\_to('pillow', 'floor'), 2. pick\_up('pillow', 'floor'), 3. move\_to('couch'), 4. put('pillow', 'couch'), 5. done().} It is assumed that the agent has a set of primitive skills or a low-level policy that allows it to execute these commands.

In this study, we focus on tasks that can be accomplished by a home robot (embodied agent) capable of navigating a household and interacting with its surroundings using a gripper, utilizing a predefined set of object-centric skills $S$. Each skill is identified by a name $n_s$ and accepts a set of arguments ${arg_i}$, which consist of open-vocabulary descriptions representing objects $o$ to be interacted with or locations $l$ where the objects may be found. In our primary experiments, the robot utilizes three distinct skills:
\begin{enumerate}
    \item $move\_to(o, l)$: Navigates the robot to object $o$ at location $l$.
    \item $pick\_up(o, l)$: Directs the robot to pick up object $o$ at location $l$.
    \item $put(o, l)$: Guides the robot to place the held object $o$ at the target location $l$.
\end{enumerate}
When an object's location is unknown, we use the placeholder ``unspecified''. This instructs the agent to employ its own heuristics (e.g., finding the nearest instance or using a tracked object) to resolve the location.

The selection of three skills mirrors the practical limitations found in many real-world robotic platforms. However, it is important to emphasize that HELP is designed to be agnostic to both the size and complexity of the underlying skill set $S$. We chose these actions specifically to validate the core functionality of our method.

To further demonstrate that HELP generalizes effectively to larger and more diverse skill sets, we present experiments on the ALFRED dataset in Section V.F, which involves a broader range of low-level skills.

\subsection{HLP Prompt Design}
The HLP processes natural language commands by decomposing the original instruction into a sequence of elementary natural language sub-tasks suitable for the LLP. Consequently, we use natural language to construct its prompts. When environmental feedback is required for task completion, a comma-separated list of relevant objects retrieved from the environment is appended to the task description within the prompt (see Figure~\ref{fig:method_overview}).

To enhance the model's adaptability and creativity, we provide only six in-context examples in the short-term memory ($M_e^H$), covering three distinct scenarios:
\begin{enumerate}
    \item A query containing a common object name that requires disambiguation.
    \item A query that can be passed directly to the LLP without modification.
    \item A complex query containing multiple objects or conjunction commands (e.g., using ``and'' or commas) that must be decomposed into elementary sub-tasks.
\end{enumerate}
Furthermore, because the long-term memory ($M_m$, e.g., a multimodal map or a segmentation model) may return multiple instances of the same object type (distinguished by an index), we also include examples featuring varying numbers of objects to improve robustness. The prompt for HLP is provided in Appendix, Figure~\ref{fig:hlp_prompt_appendix}.

\subsection{LLP Prompt Design}
\label{subsec:prompt-design}
We use the following principles to construct the low-level contextual examples.

\textbf{Structured Output}: Actions are presented as a numbered sequence, which improves the traceability of individual steps.

\textbf{Pseudocode Format}: Function names and arguments are formatted in a Python-like pseudocode style. This offers two key advantages: 1) immutability -- function names in code are fixed, which prevents the use of synonyms (e.g., ensuring $move\_to$ is used instead of $go\_to$); 2) explicit delimiters -- arguments are enclosed in quotation marks and parentheses (e.g., $move\_to('cup', 'table')$), which provides clear contextual boundaries for the LLM and simplifies the parsing and reasoning process (see Figure~\ref{fig:method_overview}).

\textbf{Diverse Examples}: We manually assembled a substantial collection of 30 varied in-context examples. Prior to planning, we select five examples based on sentence similarity between the current task and the example tasks. These examples help the model maintain the intended output format and enhance the consistency of action sequences.

\textbf{Novel Object Utilization}: The examples intentionally feature objects that are unlikely to appear in the evaluation tasks. This forces the model to rely solely on the information provided in the current query to resolve an object's location, rather than recalling locations from the in-context examples.

The complete prompt used for the LLP is provided in Appendix, Figure~\ref{fig:llp_prompt_appendix}.

\subsection{Feedback Request Prompt Design}
The role of the feedback agent is to analyze a natural language goal and determine which object or set of objects must be retrieved from the environment and transmitted as feedback to the HLP. While this can be framed as a sequence annotation task (e.g., named entity recognition), we leverage the capabilities of a language model to handle complex linguistic nuances. For instance, the model can simplify an ambiguous query like ``toys of all colors'' to the core concept ``toys,'' which subsequently improves the recall and final quality of the open-vocabulary segmentation module by focusing on a broader category.

The agent's reasoning is guided by a short profile $I^F$ and three contextual examples provided in its short-term memory ($M_e^F$). These examples cover key scenarios: retrieving a specific object instance, retrieving all objects of a given type, and a case where no feedback is required because the instruction is already unambiguous. 

In the real-world deployment, the feedback agent queries an open-vocabulary segmentation module on the robot. This module processes visual data from the robot's sensors and returns a list of segmented objects present in the environment. The feedback agent then sends the relevant information to the HLP.
The prompt used for the feedback agent is provided in Appendix, Figure~\ref{fig:feedback_request_prompt_appendix}.

\subsection{The Valid Plan Check Prompt Design}
\label{sec:check}
Performing experiments on a real robot entails inherent risks to nearby humans and material assets. To mitigate these risks, we introduced an additional LLM-based verification agent into our pipeline for all real-robot experiments. The role of this agent is to assess the feasibility and safety of a natural language task for a home robot equipped with a specific set of primitive navigation and manipulation skills, including $move\_to(object, location)$, $pick\_up(object, location)$, and $put(object, location)$. 

For each user command, the agent outputs a binary verdict: \textbf{Feasible} or \textbf{Not feasible}. If a command is deemed feasible, it is passed to the HLP for further processing. If a command is deemed not feasible, its execution is immediately halted. The user is notified of the issue, and the system waits for the next instruction. The complete prompt for this verification agent is provided in Appendix, Figure~\ref{fig:valid_check_prompt_appendix}.

This implementation demonstrates a key advantage of HELP framework: its modularity allows for the straightforward integration of new, specialized agents (like this safety checker) into the pipeline to enhance robustness and safety for real-world deployment.

\section{Experiments}\label{sec:experiments}
In this section, we first present the research questions that frame our evaluation and then outline the experimental methodology, including the datasets, evaluation metrics, and baseline comparisons.
Our evaluation aims to answer the following research questions (RQs):
\begin{itemize}
    \item \textbf{RQ1:} Is a hierarchical planning approach necessary? (Section~\ref{subsec:hier-planning})
    \item \textbf{RQ2:} How does the choice of LLM affect the performance of the low-level planner? (Section~\ref{subsec:diffmodel})
    \item \textbf{RQ3:} How does the way examples are chosen impact the performance of the low-level planner?(Section~\ref{subsec:diffprompts})
    \item \textbf{RQ4:} Does HELP generalize to different types of tasks? (Section~\ref{subsec:alfred-test})
    \item \textbf{RQ5:} Is the Valid Check module capable of distinguishing between feasible and unfeasible embodied tasks?(Section~\ref{subsec:fesibility-test})
    \item \textbf{RQ6:} Is HELP applicable to real-world settings? (Section~\ref{subsec:robot-test})
\end{itemize}

\subsection{Datasets}
\label{sec:dataset}
To evaluate our approach, we utilized common household mobile object manipulation tasks. The primary task involves moving objects from one location to another, encompassing both scenarios where the initial position is specified and must be incorporated into the plan, and scenarios where it is left unspecified. For this purpose, we generated a dedicated dataset using natural language templates, which was employed in our main experiments. The dataset comprises three distinct task types: the simple task of picking up a specified object (\textbf{Pick}), the task of picking up an object and moving it to a specified location (\textbf{Pick\&Place}), and the more complex sequential execution of two distinct pick-and-place tasks (\textbf{Pick\&Place2}). A total of 30 natural language instruction templates, 38 unique object names, and 8 possible location names were used in the dataset construction. Critically, the object names include both simple labels, such as ``apple'', and descriptions requiring property-based reasoning, such as ``green apple'', thereby ensuring the planning process must account for specific object attributes.

\subsection{Evaluation Metrics}
\begin{figure}[h]
    \centering
    \includegraphics[width=\linewidth]{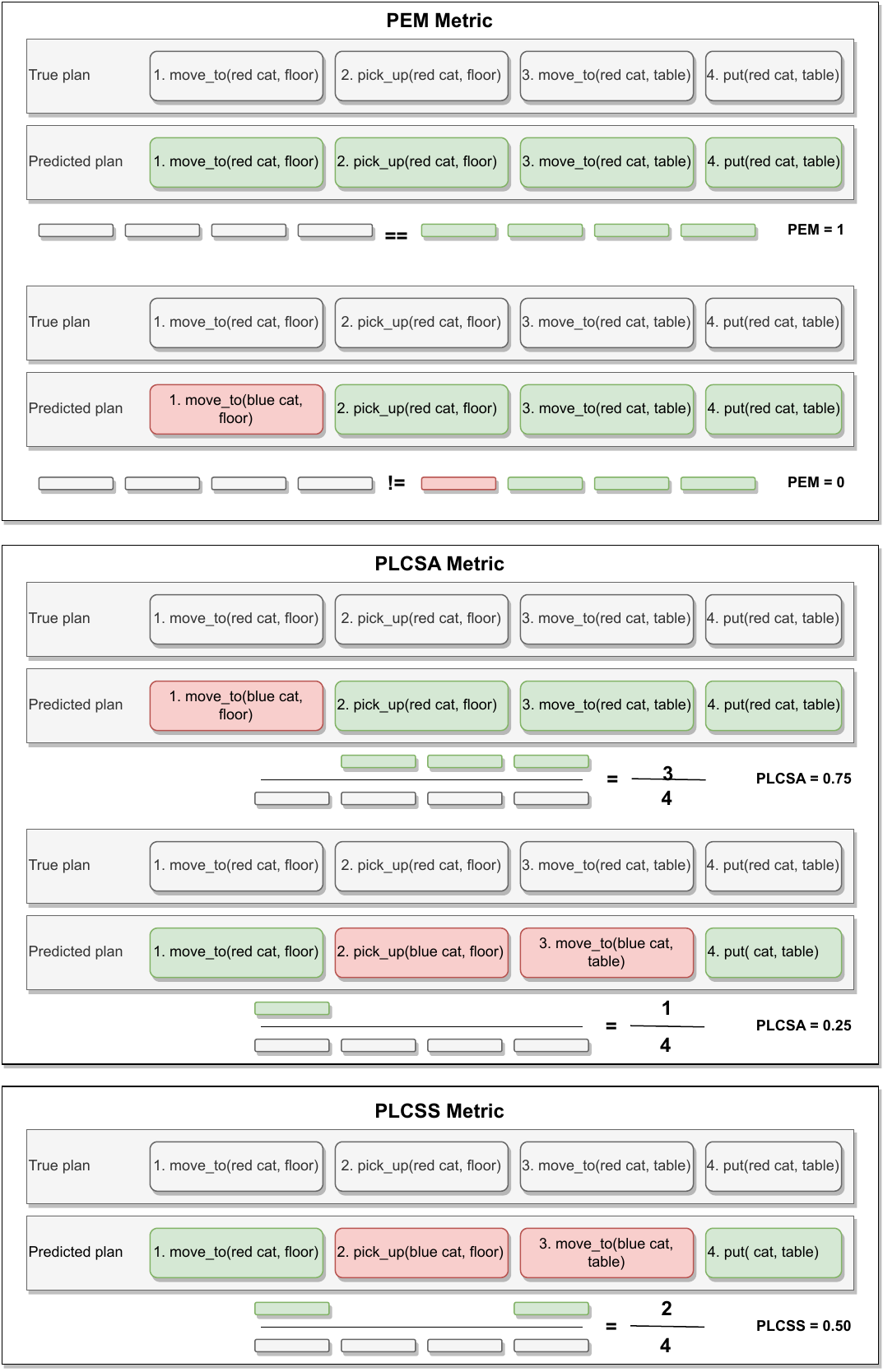}
    \caption{Illustration of plan evaluation metrics — PEM, PLCSA, and PLCSS — showing how each scores a predicted plan against the ground truth, highlighting sensitivity to action type vs. arguments and sequence continuity}
    \label{fig:metrics_explained}
    \vspace{-18pt}
\end{figure}

We evaluate the quality of the generated plans using three core metrics: Exact Match, Longest Common Subarray, and Longest Common Subsequence. \textbf{Exact Match (EM)} measures strict accuracy by determining if a predicted plan is identical to the ground truth plan in both the sequence of actions and their arguments. The \textbf{Longest Common Subarray (LCSA)} metric calculates the length of the longest continuous block of correctly predicted actions, thereby assessing the local coherence and uninterrupted overlap with the ground truth. In contrast, the \textbf{Longest Common Subsequence (LCSS)} metric evaluates the longest sequence of correct actions even if they are non-consecutive, providing a robust measure of overall similarity that is tolerant of interruptions in the sequence. These concepts are illustrated in Figure~\ref{fig:metrics_explained}. For each metric, we employ two distinct variations: one that considers only the action type (denoted \textbf{A}), and another that requires both the action and its arguments to be correct (denoted \textbf{P}). The action-only variant is instrumental, as accurately predicting the action sequence constitutes the fundamental planning challenge; errors in argument prediction can often be resolved in a subsequent grounding step.

It is important to note that our evaluation framework deliberately avoids execution-oriented metrics such as Success Rate (SR), Goal-Conditioned Success Rate (GSR), and their path weighted versions. While common in the field, these metrics measure the outcome of plan execution, which is confounded by factors beyond planning, such as simulator physics or robotic control failures. Consequently, a perfect plan (EM = 1) can still result in execution failure (SR = 0 and GSR < 1). Our chosen metrics are specifically designed to isolate and directly assess the planning capability of the HELP approach, independent of downstream execution errors. Under the ideal assumption of perfect action execution, EM would be directly equivalent to SR.

\subsection{RQ1: The Necessity of Hierarchical Planning}
\label{subsec:hier-planning}
In real-world scenarios, a person can readily issue a complex instruction comprising a long sequence of steps (e.g., ``do this, then do that, and then do this''). To evaluate the planner's ability for handling long-horizon tasks, we designed experiments on plans of varying lengths, from 2 to 16 steps. We generated 200 plans for each length category based on the dataset described in Section~\ref{sec:dataset}. For in-context learning, we utilized examples of plans ranging from 2 to 8 steps.
For baseline comparison, we evaluated four approaches: 1) ChatGPT~\cite{chat_gpt_robotics}, prompted with a formal guideline specifying the agent’s role, skills, and structured input–output format; 2) InnerMonologue~\cite{huang2022inner}, which integrates scene feedback to improve task execution; 3) ProgPrompt~\cite{singh2023progprompt}, which handles execution errors by requesting plans in code format with corresponding assertions; and 4) a standalone Low-Level Planner (LLP). For ChatGPT, we used GPT-4.1~\cite{achiam2023gpt} (gpt-4.1-2025-04-14) and GPT-3.5-Turbo~\cite{brown2020language} (gpt-3.5-turbo-0125), while for the other approaches we employed Vicuna models~\cite{vicuna2023}, which enable evaluation of the impact of different model sizes of 7B and 13B.

\subsection{RQ2: Impact of Model Selection on the LLP}
\label{subsec:diffmodel}
To systematically evaluate the impact of underlying model selection on the LLP performance, we designed a comparative experiment focusing on mid-sized, open-source large language models. 
The evaluation suite included LLaMA2~\cite{touvron2023llama}, Alpaca~\cite{alpaca}, DeepSeek-r1~\cite{guo2025deepseek}, Qwen2~\cite{bai2023qwen}, Gemma-2~\cite{team2024gemma}, Mistral~\cite{jiang2023mistral7b}, and Vicuna~\cite{vicuna2023}.
Each model was used in its standard, pre-trained form without any task-specific fine-tuning or additional optimizations to ensure a fair comparison of their inherent capabilities. The evaluation was conducted on the comprehensive dataset detailed in Section~\ref{sec:dataset}, which encompasses a wide range of planning scenarios.

\subsection{RQ3: Impact of examples selection on the HELP}
\label{subsec:diffprompts}
This experiment evaluated the impact in-context example selection on plan prediction quality for tasks of varying lengths. 
The evaluation was conducted under two distinct conditions: first, using a fixed, manually curated set of few-shot examples (Figure~\ref{fig:fixed_fewshots}), and second, employing a dynamic selection of examples based on sentence similarity to the input query (Figure~\ref{fig:variable_fewshots}).

\subsection{RQ4: Generalization of HELP to Different Tasks}
\label{subsec:alfred-test}
To demonstrate that our HELP approach generalizes to other tasks requiring a greater number of robot skills, we evaluated it on the ALFRED benchmark~\cite{alfred}. This benchmark consists of complex housekeeping tasks designed for embodied agents. 

The agent's action space in ALFRED includes eight diverse skills (GotoLocation, PickupObject, ToggleObject, PutObject, SliceObject, CleanObject, HeatObject, CoolObject).
We tested on the \textit{valid\_unseen} environment split and used examples from the \textit{valid\_seen} split for in-context learning within the prompt. For this experiment, Vicuna-7B was used as the underlying LLM for the HELP framework and baselines.

\subsection{RQ5: Feasibility module performance}
\label{subsec:fesibility-test}
To examine the ability of the Valid Check module to judge whether a command is feasible for an embodied agent, we created a subset of 200 tasks focused on feasibility assessment. Half of these tasks were intentionally infeasible by robotic manipulator (e.g., ``write a Python script'' or ``water the plants''), while the remaining 100 tasks were feasible and sampled from our dataset of plans with varying lengths.

\subsection{RQ6: Application of HELP to Real-World Robot}
\label{subsec:robot-test}
To demonstrate the applicability of the HELP framework in real-world scenarios, we deployed it on a physical embodied agent operating in a mock apartment consisting of two rooms with objects placed at random locations. We integrated HELP into the control system of a mobile robot based on the STRL architecture~\cite{mironov2023strl}. The robot is composed of a HUSKY\footnote{\url{https://clearpathrobotics.com/husky-unmanned-ground-vehicle-robot/}} mobile platform and a UR5\footnote{\url{https://www.universal-robots.com/products/ur5-robot/}} robotic arm. 
Due to the inherent risks of controlling a real robot with an LLM, we incorporated verification agent to assess the feasibility and correctness of natural language instructions before execution.

To provide environmental feedback, we utilized an open-vocabulary semantic segmentation module integrated into the STRL control system. This module returns a list of segmented objects in the environment given a language query. The robot was tasked with performing pick-and-place operations on different objects randomly distributed in the apartment. All low-level skills, including object localization, navigation, and manipulation, were handled by the STRL control system. We conducted 15 evaluation episodes to rigorously test the system.

\section{RESULTS}
\label{sec:results}
In this section, we present the results obtained from our experiments. We then provide a detailed analysis of these results, discussing their implications and how they address the research questions posed in this work.

\subsection{RQ1: The Necessity of Hierarchical Planning}
\begin{figure}[ht]
    \centering
    \includegraphics[width=\linewidth]{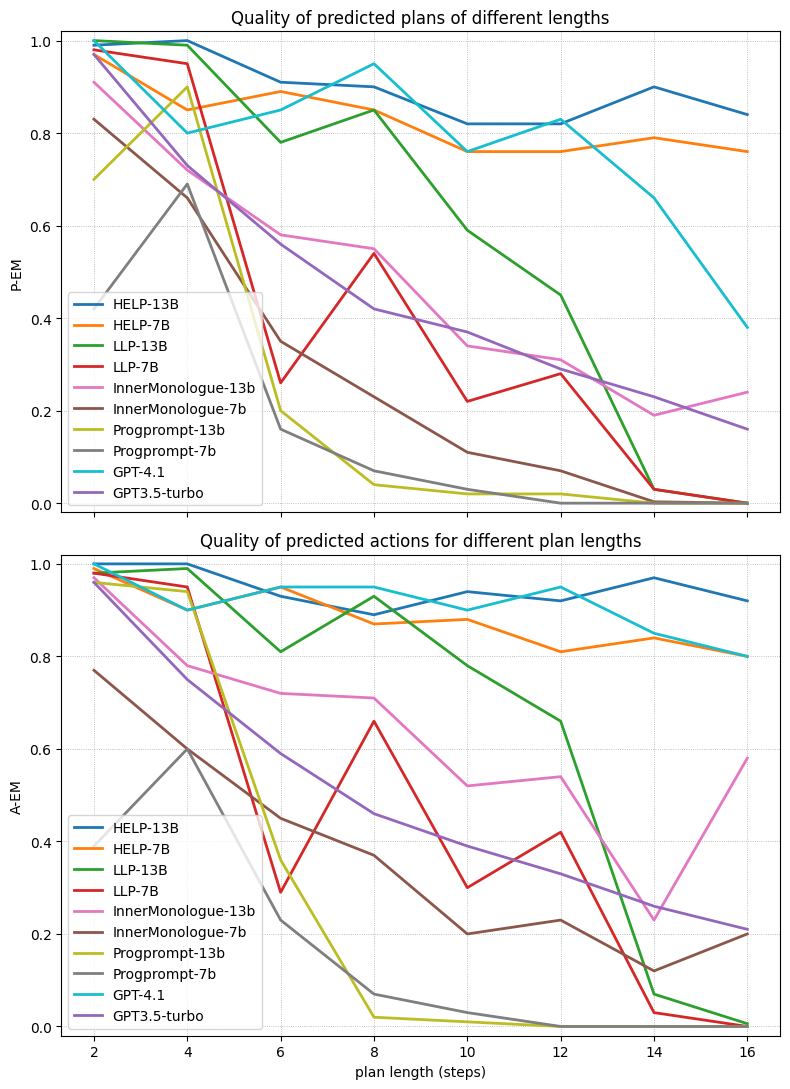}
    \caption{A comparison of the HELP, LLP, with baseline approaches, tested on plans ranging from 2 to 16 steps in length. Each length category comprised 200 plans.}
    \label{fig:models_plans_comparison}
    \vspace{-15pt}
\end{figure}

The results for plans of varying lengths are shown in Figure~\ref{fig:models_plans_comparison}. Baseline methods maintained stable performance only on short plans of up to 4 actions, typically involving interaction with a single object (e.g., ``move to, pick up, move to, put down’’).
All LLM-based planners display inconsistent behavior and have difficulty recognizing when a task is complete, especially for tasks that require just holding an object (e.g. 6,10,14 actions tasks). Even with clear instructions, they often attempt to place the object before ending the task, likely due to autoregressive plan generation from few-shot examples.
ProgPrompt required generating broad assertions before each step, which enlarged context and often confused later predictions. InnerMonologue alleviated this issue and, with scene feedback, outperformed the LLP planner, though its accuracy still declined as plan length increased.
ChatGPT with the largest closed model, GPT-4.1, was able to construct plans of up to 12 steps, but performance dropped sharply beyond that, with hallucinated object names reducing P-EM to 39\% on 16-step plans.
In contrast, our HELP approach exhibited only a slight degradation with longer horizons. Its performance remained robust and less dependent on model size, confirming the necessity of hierarchical planning.

\subsection{RQ2: Impact of Model Selection on the LLP}
The results of evaluating model selection on LLP performance are shown in Table~\ref{tab:llmmodelselection}. Among the tested models, Vicuna and Mistral achieved the best performance on plans with four-action sequences (e.g., \textit{move\_to}, \textit{pick\_up}, \textit{move\_to}, \textit{put}), a core task pattern for embodied agents. Gemma 2 (2B), while the most time-efficient, showed reduced performance. DeepSeek r1 demonstrated broad reasoning capabilities but failed to consistently maintain the required output structure, often producing unstructured answers.

\begin{table*}[ht]
    \centering
        \caption{Experimental results for Low Level Planner based on different LLMs. The datasets are: p (Pick), p\&p
        (Pick\&Place) and p\&p2 (Pick\&Place2). The prompt type $sc$ $t-s$, means that the prompt is composed in the saycan style, and the examples of successfully completed tasks were selected from the current dataset and specific to the current task. For each task type the number of in-context examples is 5. EM -- Exact Match, LCS(S; A) -- Longest Common Sub (Sequence; Array)}
    \begin{tabular}{rrcccrrrrrrr}
    \toprule
        \multicolumn{1}{c}{\multirow{2}{*}{Model}} & \multicolumn{1}{c}{\multirow{2}{*}{Size}}
        & \multirow{2}{*}{Dataset}
        & \multirow{2}{*}{N.Tasks}
        & \multirow{2}{*}{Prompt}
        & \multicolumn{3}{c}{Plan}
        & \multicolumn{3}{c}{Actions}
        & \multicolumn{1}{c}{\multirow{2}{*}{Time$\pm$Std}}
        \\ \cline{6-11}
        \multicolumn{1}{c}{} & \multicolumn{1}{c}{} & & & & \multicolumn{1}{c}{EM} & \multicolumn{1}{c}{LCSS}
        & \multicolumn{1}{c}{LCSA}
        & \multicolumn{1}{c}{EM}
        & \multicolumn{1}{c}{LCSS}
        & \multicolumn{1}{c}{LCSA}
        & \multicolumn{1}{c}{}
        \\
        \midrule
        LLaMA2 & 7B & p & 200 & sc
        t-s & \textbf{0.88} & \textbf{0.94} & \textbf{0.94} & 1.00 & 1.00 & 1.00 & 19.00$\pm$0.66 \\
        Alpaca & 7B & p & 200 & sc
        t-s & 0.81 & 0.88 & 0.88
        & 1.00 & 1.00 & 1.00
        & 5.67$\pm$0.42 \\
        DeepSeek r1 & 8B & p & 200 & sc
        t-s & 0.02 & 0.04 & 0.04
        & 0.03 & 0.08 & 0.08 
        & 14.13$\pm$3.78  \\
        Qwen 2 & 0.5B & p & 200 & sc
        t-s & 0.20 & 0.58 & 0.58
        & 0.27 & 0.89 & 0.89 
        & 5.69$\pm$1.23  \\
        Gemma 2 & 2B & p & 200 & sc
        t-s & 0.59 & 0.79 & 0.79
        & 0.60 & 0.81 & 0.81 
        & \textbf{2.31$\pm$0.70} \\
        Mistral & 7B & p & 200 & sc
        t-s & 0.81 & 0,87 & 0.87
        & 0.84 & 0.92 & 0.92 
        & 3.34$\pm$1.69  \\
        Vicuna & 7B & p & 200 & sc
        t-s & 0.79 & 0.85 & 0.85
        & \textbf{1.00} & \textbf{1.00} & \textbf{1.00} & 3.54$\pm$0.11 \\
        \midrule
        LLaMA2 & 7B & p\&p & 200 & sc
        t-s & 0.66 & 0.91 & 0.90
        & \textbf{1.00} & 1.00 & 1.00
        & 20.50$\pm$1.28 \\
        Alpaca & 7B & p\&p & 200 & sc
        t-s & 0.69 & 0.91 & 0.89
        & 0.97 & 0.99 & 0.99
        & 9.31$\pm$0.78 \\
        Qwen 2 & 0.5B & p\&p & 200 & sc
        t-s & 0.28 & 0.48 & 0.46
        & 0.44 & 0.71 & 0.66 
        & 4.50$\pm$0.93  \\
        Gemma 2 & 2B & p\&p & 200 & sc
        t-s & 0.55 & 0.61 & 0.60
        & 0.61 & 0.71 & 0.70 
        & \textbf{2.52$\pm$0.74}  \\
        Mistral & 7B & p\&p & 200 & sc
        t-s & 0.87 & 0.95 & 0.94
        & 0.94 & 0.98 & 0.98 
        & 3.28$\pm$0.56  \\
        Vicuna & 7B & p\&p & 200 & sc
        t-s & \textbf{0.88} & \textbf{0.96} & \textbf{0.96} & 0.99 & \textbf{1.00}
        & \textbf{1.00}
        & 3.76$\pm$0.30 \\
        \midrule
        LLaMA2 & 7B & p\&p2 & 200 & sc
        t-s & 0.23 & 0.79 & 0.69
        & 0.60 & 0.93 & 0.92
        & 21.83$\pm$1.28 \\
        Alpaca & 7B & p\&p2 & 200 & sc
        t-s & 0.27 & 0.81 & 0.73
        & 0.55 & 0.92 & 0.92
        & 16.98$\pm$0.70 \\
        Qwen 2 & 0.5B & p\&p2 & 200 & sc
        t-s & 0.01 & 0.26 & 0.21
        & 0.10 & 0.61 & 0.46 
        & 4.17$\pm$1.81  \\
        Gemma 2 & 2B & p\&p2 & 200 & sc
        t-s & 0.10 & 0.25 & 0.21
        & 0.13 & 0.46 & 0.33 
        & \textbf{3.22$\pm$0.79}  \\
        Mistral & 7B & p\&p2 & 200 & sc
        t-s & \textbf{0.74} & \textbf{0.91} & \textbf{0.88}
        & \textbf{0.83} &\textbf{ 0.96} & \textbf{0.93} 
        & 3.80$\pm$0.44  \\
        Vicuna & 7B & p\&p2 & 200 & sc
        t-s & 0.34 & 0.84 & 0.78 & 0.60
        & 0.93
        & 0.92
        & 4.34$\pm$0.11  \\ 
        \bottomrule
        \label{tab:pp}
    \end{tabular}
    \label{tab:llmmodelselection}
    \vspace{-15pt}
\end{table*}

\subsection{RQ3: Impact of examples selection on the HELP}
\begin{figure*}[ht]
    \centering
    \includegraphics[width=\linewidth]{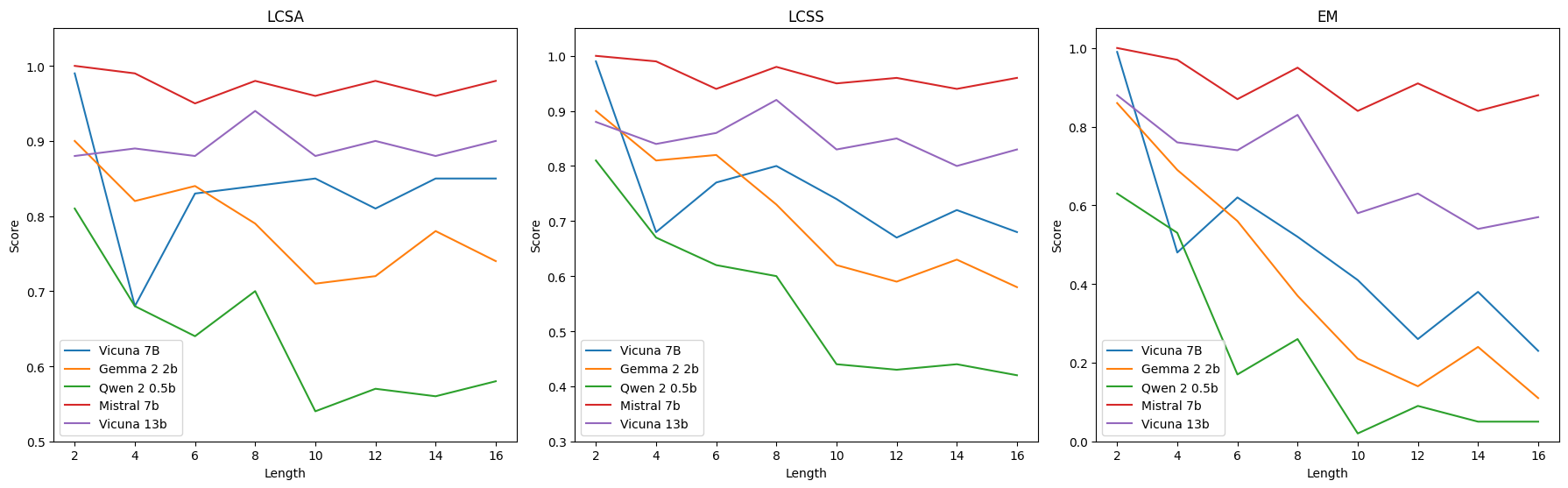}
    \caption{Compared to fixed examples set, all models show improvement across all metrics. This demonstrates that dynamically selecting relevant examples based on instruction similarity is more effective than fixed prompts.}
    \label{fig:variable_fewshots}
\end{figure*}

The results across all evaluated models under both fixed and similarity-based few-shot settings are shown in Figures~\ref{fig:fixed_fewshots} and \ref{fig:variable_fewshots}. The findings indicate that selecting examples based on similarity leads to a clear and consistent improvement in the quality of plan generation. Importantly, this effect cannot be explained by model size alone: every model, regardless of its scale, benefits from a comparable boost when similarity-based examples are used. This suggests that the choice of examples plays a critical role in enhancing HELP performance.

\begin{figure*}[ht]
    \centering
    \includegraphics[width=\linewidth]{img/fig5_fixed_fewshots.png}
    \caption{The plots show LCSA, LCSS, and EM scores versus plan length. While larger models generally perform better, all models exhibit performance degradation as task length increases}
    \label{fig:fixed_fewshots}
\end{figure*}

\subsection{RQ4: Generalization of HELP to Different Tasks}
Table~\ref{tab:alfredllm} summarizes the results of evaluating HELP across different task types. 
While InnerMonologue and ProgPrompt handle short-horizon tasks from the ALFRED dataset, their performance degrades as task complexity increases. In tasks that require more different actions, they often confuse action order, for example, attempting to slice an object before picking up the knife or removing an item from the microwave before heating it.
HELP addresses this issue by decomposing a task into subtasks and leveraging k-shot examples, which improves action sequencing. However, due to the benchmark’s strict requirement for exact matches with object names in the environment, HELP without environmental grounding achieved only a PEM score of 0.35.  
To overcome this, we introduced an environmental grounding module that retrieves existing objects by computing cosine similarity between Sentence-BERT~\cite{reimers-2019-sentence-bert} embeddings of candidate items. By comparison, InnerMonologue relies on scene descriptions, and ProgPrompt has no grounding mechanism. With the grounding module, HELP’s performance improved significantly, reaching a PEM score of 0.64 (Table~\ref{tab:alfredllm}). This highlights the critical role of perceptual grounding in executing complex tasks within unfamiliar environments.

\begin{table*}[h]
\centering
    \caption{Experimental results for the generalization of HELP on tasks from the ALFRED dataset (\textit{valid\_unseen} split).}
  \begin{tabular}{cclcccrrcrr}
\toprule
\multirow{2}{*}{Model}     & \multirow{2}{*}{Size}  & \multirow{2}{*}{Method} & \multirow{2}{*}{N.Tasks}     & \multirow{2}{*}{Grounding}    & \multicolumn{3}{c}{Plan}                                                       & \multicolumn{3}{c}{Actions}                                                    \\ \cline{6-11} 
                           &                        &                         &                         &                            & EM                       & \multicolumn{1}{c}{LCSS} & \multicolumn{1}{c}{LCSA} & EM                       & \multicolumn{1}{c}{LCSS} & \multicolumn{1}{c}{LCSA} \\ 
                           \midrule
\multicolumn{1}{r}{Vicuna} & \multicolumn{1}{r}{7B} & Alfred                  & \multicolumn{1}{r}{812} & \multicolumn{1}{r}{No} & \multicolumn{1}{r}{0.35} & 0.64                     & 0.59                     & \multicolumn{1}{r}{0.64} & 0.89                     & 0.83                     \\
\multicolumn{1}{r}{Vicuna} & \multicolumn{1}{r}{7B} & Alfred                  & \multicolumn{1}{r}{812} & \multicolumn{1}{r}{Yes} & \multicolumn{1}{r}{\textbf{0.64}} & \textbf{0.86}                     & \textbf{0.78}                     & \multicolumn{1}{r}{\textbf{0.91}} & \textbf{0.96}                     & \textbf{0.94}                     \\
\multicolumn{1}{r}{Vicuna} & \multicolumn{1}{r}{7B} & InnerMonologue          & \multicolumn{1}{r}{812} & \multicolumn{1}{r}{Yes} & \multicolumn{1}{r}{0.15}   & 0.50                       & 0.38                       & \multicolumn{1}{r}{0.21}   & 0.65                       & 0.49                       \\
\multicolumn{1}{r}{Vicuna} & \multicolumn{1}{r}{7B} & ProgPrompt              & \multicolumn{1}{r}{812} & \multicolumn{1}{r}{No} & \multicolumn{1}{r}{0.12}   & 0.26                       & 0.20                       & \multicolumn{1}{r}{0.18}   & 0.61                       & 0.49                       \\ 
\bottomrule
\end{tabular}

\label{tab:alfredllm}
\end{table*}

\subsection{RQ5: Feasibility module performance}
The Task Feasibility Estimator achieved 77\% accuracy in predicting whether a task was feasible. It reliably identified the unfeasibility of tasks such as ``write Python code'' or ``plan a day trip to a nearby city.'' However, it struggled with tasks involving assembly or object manipulation that exceed the robot’s capabilities, which are limited to pick-up actions. Examples include ``assemble a jigsaw puzzle'' or ``fold and put away the laundry,'' both of which were incorrectly marked as feasible. These results indicate that the module is effective overall, but further improvements are needed to better handle complex or ambiguous tasks.

\subsection{RQ6: Application of HELP to Real-World Robot}
The robot successfully completed 12 out of the 15 tasks, achieving an 80\% success rate. These results confirm the practical applicability of HELP for real-world embodied tasks. The experiments also revealed a critical insight: even with a perfectly generated plan from HELP, an episode can fail due to errors in low-level robot execution. For instance, failures occurred when an object could not be found, a pickup attempt failed, or a successfully grasped object was dropped. In such cases, the plan execution was interrupted without recovery. This highlights a vital direction for future research: enhancing HELP with re-planning capabilities that can incorporate real-time feedback from failed low-level actions to dynamically adjust the plan.

\begin{figure*}[ht]
    \centering
    \includegraphics[width=0.9\linewidth]{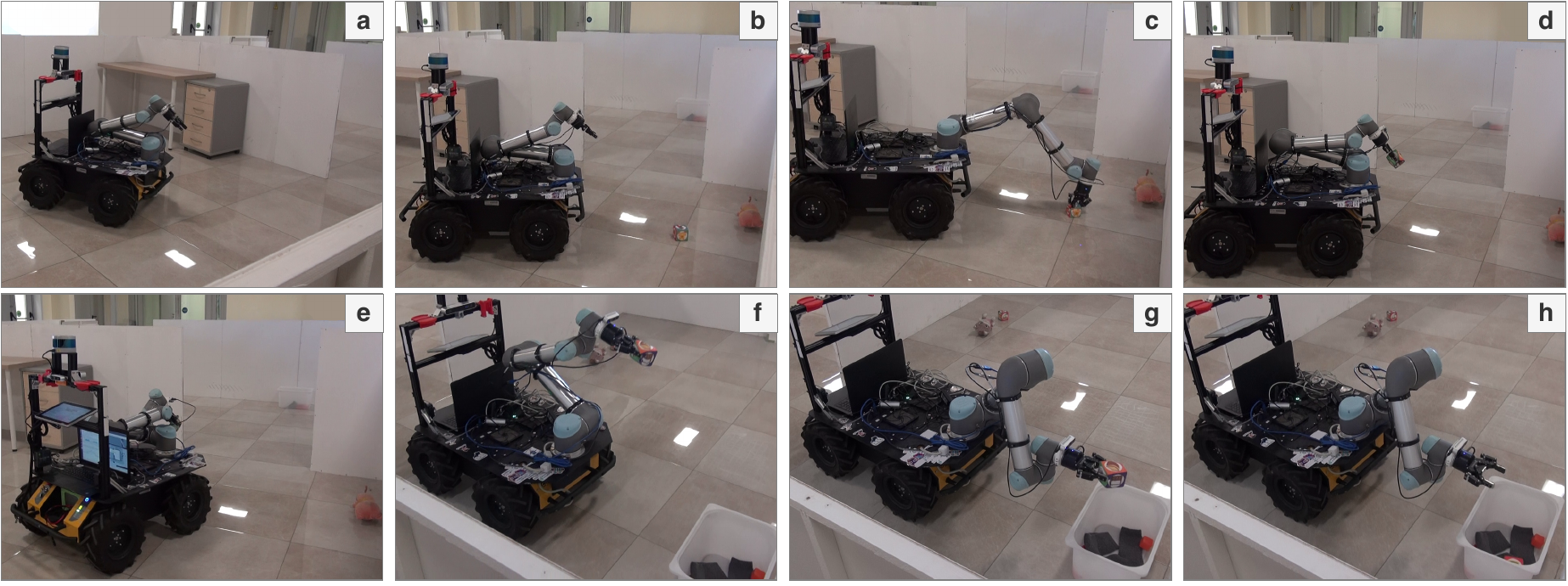}
    \caption{An example of plan execution on a real embodied agent. HELP generated the plan according to the natural language instruction ``\textit{put the toy cube in the white box}''. The plan is: 1. move\_to("toy cube", "unspecified") \textbf{(steps a-b)}, 2. pick\_up("toy cube", "unspecified") \textbf{(steps c-d)}, 3. move\_to("toy cube", "white box") \textbf{(steps e-f)}, 4. put("toy cube", "white box") \textbf{(steps g-h)}.}
    \label{fig:robot}
    \vspace{-14pt}
\end{figure*}

Figure~\ref{fig:robot} illustrates a successful execution of the natural language instruction ``\textit{put the toy cube in the white box}''. The sequence shows the robot determining the cube's location and navigating to it (a-b), picking up the cube (c-d), moving to the white box (e-f), and finally placing the cube inside it (g-h), culminating in a successful episode.

\section{DISCUSSION}
\label{sec:discussion}
Our approach is based on a large language model that operates in generation mode, and despite the use of a prompt that limits the output format, the model can potentially generate inappropriate and/or offensive output. Also, language models are prone to hallucinations and can generally produce unforeseen results, so giving them control over mechanisms that could potentially cause harm, such as a mobile robot with a robotic arm, and testing such mechanisms should be done in a regulated manner, in a specially designated area with limited access to the people involved in the experiments. It is also potentially possible to deliberately execute harmful plans on a robot with the intent to cause harm. Modern language models can be partially effective in thwarting such attempts. For example, in experiments we observed that when the word ``kitten'' was used in the wording of the instructions as a reference to the object to be moved, the model refused to generate a plan and generated text describing that it could not manipulate living creatures. Changing the word to ``toy kitten'' allows LLM to start planning. However, to reduce these risks, we use a special prompt that skips tasks involving the manipulation of inanimate objects for household tasks to the next stage before moving on to planning. However, we rely on the capabilities of the model itself, which cannot guarantee an absolute result.

The limitations of our approach include, first, working only with instructions in English, which limits access to the results for people who do not speak that language. This is mainly due to the availability of open source LLMs for English compared to LLMs for other languages. Another limitation is the set of actions used during planning, which is related to the capabilities of the real robot on which the experiments were performed. This limitation is not related to the architectural features of HELP, and the list of possible actions can be freely extended, assuming the potential possibility of these actions taking into account the environment and the real embodied agent. We have also limited the scope of the considered tasks to household tasks, which is not a limitation of the architectures, but related to the specific application of a real robot. Therefore, the application of HELP can be generalized to other domains. Although we have only tested our approach on a specific embodied agent, this is potentially not a limitation, since HELP is agnostic to the realization of the embodied agent, provided that the planning uses the actions available to the agent.

\section{CONCLUSION}
\label{sec:conclusion}
We presented HELP, a hierarchical approach to robot task planning using LLM-based agents. This architecture leverages the capabilities of LLMs by decomposing complex instructions into manageable sub-goals, leading to more accurate and reliable planning. Our method facilitates the use of medium-sized, open-source models that can run directly on a single GPU within an embodied agent, thereby increasing autonomy and eliminating dependence on proprietary APIs. The framework is inherently flexible, allowing for the integration of additional specialized LLM agents to extend its capabilities, such as for safety verification or environmental feedback. We evaluated the proposed approach on household tasks of varying complexity. Furthermore, we tested HELP on the ALFRED benchmark to demonstrate its generalization to more complex tasks requiring a broader set of skills. To prove the practical applicability of HELP in real-world scenarios, we integrated it into the control system of a real embodied agent, a mobile platform equipped with a robotic arm, and demonstrated that it can successfully function as a task planner. We hope that the HELP framework will advance the application of LLM-based planners for real-world robotic systems.

\bibliographystyle{IEEEtran}
\bibliography{IEEEexample}
\newpage
\onecolumn
\section*{APPENDIX A -- PROMPT DESIGN}

This appendix provides the complete prompt templates employed in the HELP architecture. We include representative examples for all core components of the system: the High-Level Planner, the Low-Level Planner, the environment feedback query agent, and the task feasibility verification module. These prompts are instrumental in shaping the behavior of the LLM-based agents and ensuring consistent, structured plan generation throughout the planning pipeline.

\begin{figure*}[h]
    \centering
    \includegraphics[width=0.8\linewidth]{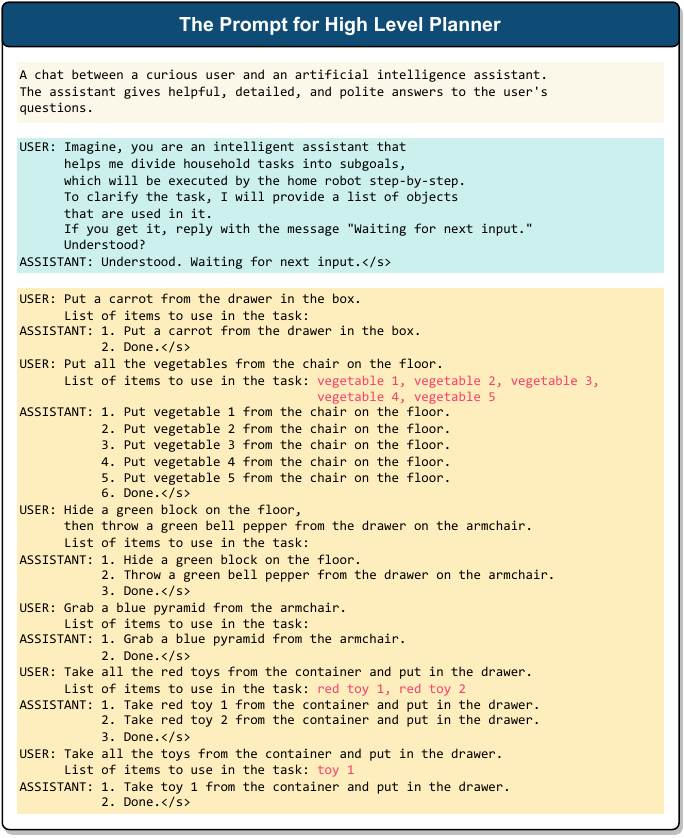}
    \caption{High Level Planner prompt example.}
    \label{fig:hlp_prompt_appendix}
\end{figure*}

\begin{figure*}[ht]
    \centering
    \includegraphics[width=0.9\linewidth]{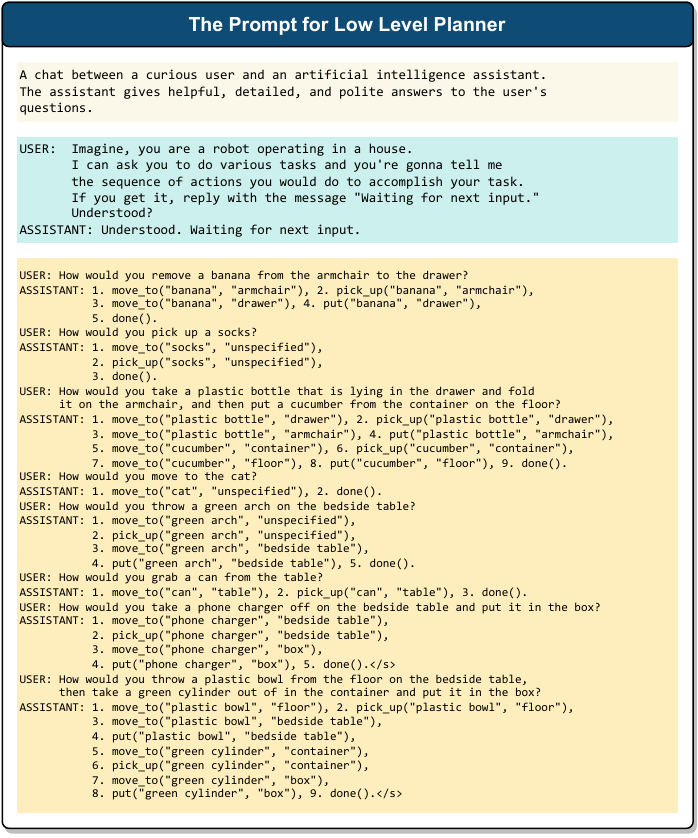}
    \caption{Low Level Planner prompt example.}
    \label{fig:llp_prompt_appendix}
\end{figure*}

\begin{figure*}[ht]
    \centering
    \includegraphics[width=0.75\linewidth]{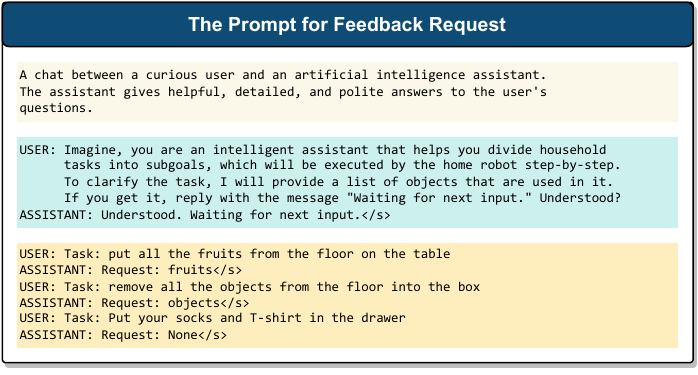}
    \caption{Feedback request prompt example.}
    \label{fig:feedback_request_prompt_appendix}
\end{figure*}
\begin{figure*}[ht]
    \centering
    \includegraphics[width=0.75\linewidth]{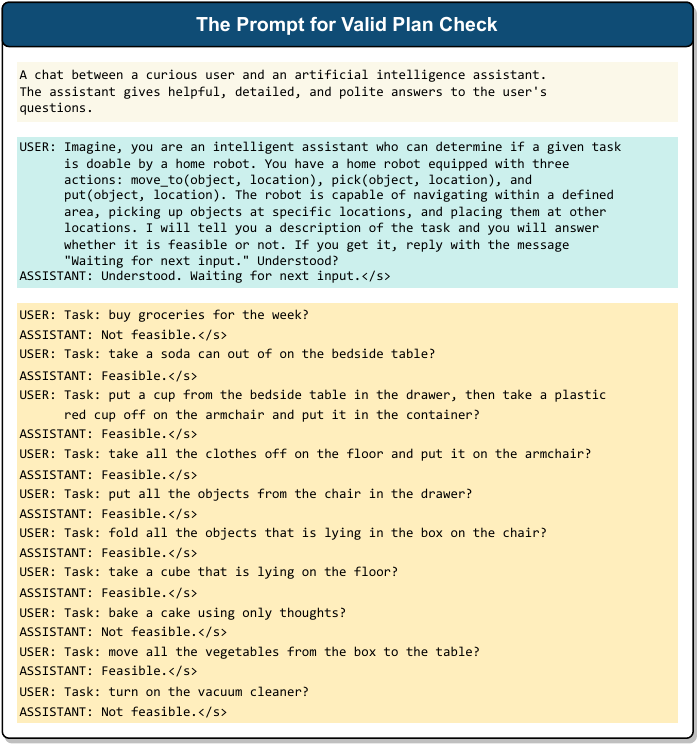}
    \caption{Valid plan check prompt example.}
    \label{fig:valid_check_prompt_appendix}
\end{figure*}

\end{document}